\pdfoutput=1
\documentclass[11pt]{article}

\usepackage[final]{ACL2023}

\usepackage{amsmath}
\usepackage{times}
\usepackage{latexsym}
\usepackage[capitalize]{cleveref}

\usepackage[T1]{fontenc}
\usepackage[utf8]{inputenc}

\usepackage{microtype}

\usepackage{inconsolata}

\usepackage{color,soul}
\usepackage{graphics}
\usepackage{graphicx}
\usepackage{tikz}
\usetikzlibrary{positioning, shadows, shapes, arrows}

\usepackage{subcaption}
\newcommand{\rl}[1]{\textcolor{teal}{}}
\newcommand{\ac}[1]{\textcolor{purple}{}}
\newcommand{\yg}[1]{\textcolor{green}{}}

\title{Diversity Over Quantity: A Lesson From Few Shot Relation Classification}

\author{ Amir DN Cohen$^{1}$\hspace{1em} Shauli Ravfogel$^{1}$\hspace{1em} Shaltiel Shmidman$^{3}$\hspace{1em}  Yoav Goldberg$^{1,2}$\\
    $^1$Bar-Ilan University, Ramat-Gan, Israel\\
    $^2$Allen Institute for AI, Seattle, WA\\
    $^3$DICTA, Jerusalem, Israel\\
    {\small\tt amirdnc@gmail.com,  shauli.ravfogel@gmail.com, shaltiel@dicta.org.il, ranlevy@amazon.com, yoav.goldberg@gmail}
}
\begin{document}
\maketitle
% \begin{abstract}
% What makes few-shot relation classification models (FSRC) effective? is it the sheer amount of data they are trained on? 
% Or can data-lean models achieve similar performance by carefully selecting the training data? 
% In this work, we demonstrate that in the context of few-shot relation classification, the diversity of the training data is far more important than its size: 
% we train models with a fixed amount of data but with a varying number of relations. 
% We find that few-shot performance on relations that were unseen 
% at training time substantially improves when relation type diversity increases.
% We demonstrate this on a new few-shot dataset we collected for this task and several prominent FSRC datasets. These findings suggest that targeted data curation can help alleviate the data hunger of models.
% \end{abstract}
\begin{abstract}
In few-shot relation classification (FSRC), models must generalize to novel relations with only a few labeled examples. While much of the recent progress in NLP has focused on scaling data size, we argue that \textbf{diversity} in relation types is more crucial for FSRC performance. In this work, we demonstrate that training on a diverse set of relations significantly enhances a model’s ability to generalize to unseen relations, even when the overall dataset size remains fixed.

We introduce REBEL-FS, a new FSRC benchmark that incorporates an order of magnitude more relation types than existing datasets. Through systematic experiments, we show that increasing the diversity of relation types in the training data leads to consistent gains in performance across various few-shot learning scenarios, including high-negative settings. Our findings challenge the common assumption that more data alone leads to better performance and suggest that \textbf{targeted data curation} focused on diversity can substantially reduce the need for large-scale datasets in FSRC.
\end{abstract}

\section{Introduction}
\begin{figure}[h]
    \centering
    \begin{tikzpicture}
        % Define styles
        \tikzstyle{box} = [rectangle, draw=black, fill=blue!50, rounded corners, inner sep=5pt, text width=7.5cm]
        \tikzstyle{arrow} = [thick,->,>=stealth]
        
        % Support Set
        \node[box] (support) {
            \small
            \textbf{Support Set:}
            \begin{itemize}
                \item \textit{``\underline{Einstein} was born in \underline{Ulm}.''}
                
                \textbf{Relation:} \textit{place of birth} \textbf{Entities:} \underline{Einstein}, \underline{Ulm}
                
                \item \textit{``\underline{Apple} founded by \underline{Jobs}.''}
                
                \textbf{Relation:} \textit{founded by} \textbf{Entities:} \underline{Apple}, \underline{Jobs}
            \end{itemize}
        };
        
        % Query Examples
        \node[box, below=0.1cm of support] (query) {
            \small
            \textbf{Query Example:}
 \textit{``\underline{Curie} was born in \underline{Warsaw}.''}
                
                \textbf{Task:} Assign \textit{place of birth} \quad \textbf{Entities:} \underline{Curie}, \underline{Warsaw}
        };
        
        % NOTA Example
        \node[box, below=0.1cm of query] (nota) {
            \small
            \textbf{Query Example (NOTA):}
            \textit{``\underline{Alice} reads \underline{books}.''}
                
                \textbf{Task:} Assign \textbf{NOTA} \quad \textbf{Entities:} \underline{Alice}, \underline{books}
        };
        
    \end{tikzpicture}
    \caption{Illustration of the few-shot relation classification setting. The model uses a \textbf{Support Set} with few examples to classify relations in \textbf{Query Examples}, assigning the correct relation type or ``non-of-the-above'' \textbf{NOTA} if it doesn't match known relations.}
    \label{fig:fsrc-example}
\end{figure}

Few-shot relation classification (FSRC) is a pivotal NLP task \cite{gao-etal-2019-fewrel, Sabo2021, han-etal-2018-fewrel}, where models are required to classify relationships between entities with only a handful of labeled examples as shown in Figure \ref{fig:fsrc-example}. This task is particularly challenging due to the long-tail distribution of relations in real-world scenarios. While traditional benchmarks capture frequent relations, many specialized domains—such as law, medicine, and economics—rely on recognizing relations that are either rare or completely novel. For instance, relations like \textit{``acquired by''} (economics), \textit{``gene encodes protein''} (biology), or \textit{``disease treated by''} (medicine) are underrepresented in most datasets. Consequently, effective FSRC models must be able to generalize from limited examples of novel relations, making generalization performance crucial in real-world applications.

The dominant approach to improving few-shot performance in FSRC has been scaling data quantity. For instance, large language models (LLMs) such as LLAMA and T5 have demonstrated impressive performance across many NLP tasks, largely due to being trained on massive corpora of diverse text data  \cite{Touvron2024,Raffel2019, Zhou, Li2023}. However, while data scaling has yielded gains in performance, it often neglects an equally important factor: data diversity. Models trained on large but homogeneous datasets, despite their size, often fail to generalize effectively to novel relations. In the few-shot setting, where relations encountered at test time are likely to differ from those in the training data, a lack of relation diversity can severely limit the model's generalization capabilities.

In this work, we argue that \textbf{relation diversity}, rather than sheer data quantity, is the key to effective few-shot generalization. We hypothesize that training models on a diverse range of relation types exposes them to a wider variety of linguistic patterns, entity structures, and semantic relationships. This broader exposure equips models to recognize subtle variations in unseen relations, enabling better generalization. Specifically, diversity helps the model learn to capture differences in argument structures, syntactic patterns, and surface forms that are associated with different relations. %For example, a model trained on a dataset rich in relations between people and organizations (e.g., \textit{``employed by''}, \textit{``founded by''}) is more likely to generalize to novel but related relations such as \textit{``mentored by''}.

To test this hypothesis, we introduce \textbf{REBEL-FS}, a novel few-shot relation classification dataset that prioritizes diversity over data quantity. Unlike existing FSRC benchmarks, REBEL-FS covers an order of magnitude more relation types, with over 900 different relations represented. This allows us to systematically study the effects of relation diversity on few-shot generalization. We conduct controlled experiments where we fix the total number of training examples but vary the number of relation types, ranging from 10 to 400. Our findings demonstrate that increasing relation diversity significantly improves model performance across multiple FSRC datasets.

Our results show that when models are trained on more diverse relation types, the accuracy on unseen relations improves in F1 by up to \textbf{13\%} compared to models trained on a smaller set of relations, even with the same total number of examples. For instance, in one experiment, models trained on 400 relation types achieved an F1 score of \textbf{91.3\%}, compared to only \textbf{80.9\%} for models trained on 29 relation types, even though both models were trained with 100,000 examples. This suggests that relation diversity has a compounding effect on generalization, allowing the model to more effectively distinguish between positive and negative examples in high-negative settings.  We evaluated these results across multiple well-established FSRC datasets, including \textbf{TACRED-FS} \cite{Sabo2021}, \textbf{CORE} \cite{Borchert2023}, and \textbf{FewRel 2.0} \cite{gao-etal-2019-fewrel}, confirming that increased relation diversity consistently leads to better performance across different domains.

Moreover, we explore the impact of relation diversity in high-negative scenarios, where most examples are negative, a common real-world condition. Our experiments reveal that in such challenging settings, models trained on diverse relations can improve F1 scores by up to \textbf{157\%}, particularly when faced with 99\% negative examples. This indicates that relation diversity not only aids generalization to novel relations but also enhances the model's robustness in highly imbalanced datasets, where distinguishing relevant relations is inherently more difficult.

% In summary, our contributions are as follows:
% \begin{itemize}
%     \item We introduce REBEL-FS, a new FSRC dataset that covers a wider range of relation types than any previous benchmark, providing a comprehensive testbed for studying few-shot generalization.
%     \item We empirically demonstrate that relation diversity, rather than sheer data size, plays a crucial role in improving FSRC performance, especially in few-shot and high-negative settings.
%     \item We present a set of practical insights into data curation strategies for FSRC, suggesting that models can achieve competitive performance with far less data if the training set is carefully curated for diversity.
% \end{itemize}

The remainder of this paper is structured as follows: Section \ref{sec:related} reviews related work in FSRC and dataset diversity. Section \ref{sec:diversity} farther explain the diversity hypothesis. Section \ref{sec:comp-datasets} details the construction of REBEL-FS. In Section \ref{sec:expiriemts}, we present our experimental results and analysis, and finally, we summerize our findings in Section \ref{sec:conclusion}.

\section{Background and Related Work} \label{sec:related}
We begin by reviewing existing datasets and approaches in FSRC, highlighting the limitations addressed by our research. We then examine the broader importance of data diversity in other works.

\subsection{Few-Shot Relation Classification} 
\subsubsection{Background}

Relation classification is a fundamental task in natural language processing (NLP) that involves identifying the semantic relationship between pairs of entities within a text \citep{Hendrickx2010, Zhang, Cohen2020a}. Traditional approaches rely on supervised learning methods that require large amounts of annotated data. However, acquiring such extensive labeled datasets is often impractical due to the high cost and effort of manual annotation.

To address this challenge, \emph{few-shot learning} has been introduced to relation classification \citep{sabo-etal-2021-revisiting, han-etal-2018-fewrel, BaldiniSoares2019, gao-etal-2019-fewrel}. Few-shot relation classification aims to enable models to recognize new relation types given only a small number of labeled examples. This paradigm is crucial for real-world applications where new relations emerge frequently and labeled data is scarce.

Despite rapid advancements, models soon achieved near-human or superhuman performance on early benchmarks, indicating that they did not fully capture the complexities of real-world relation extraction tasks. To address this, more realistic and challenging datasets incorporating \textbf{None-of-the-Above (NOTA)} examples were introduced \citep{gao-etal-2019-fewrel, sabo-etal-2021-revisiting}. In these scenarios, the model must not only choose between a few target relations but also determine if the query instance does not correspond to any of the relations in the support set, adding an additional layer of complexity that reflects more realistic conditions where unseen relations may appear.
\subsubsection{Problem Formulation} \label{sssec:formulation}

In the few-shot relation classification setting, the task is typically formulated as an $M$-way $K$-shot classification problem. Each episode consists of:

\begin{itemize}
    \item A \textbf{support set} $\mathcal{S} = \left\{ (x_i, y_i) \right\}_{i=1}^{M \times K}$, where $x_i$ is an instance containing a pair of entities in a text, and $y_i \in \mathcal{R}$ is the relation label. The support set includes $K$ examples for each of the $M$ target relation types. Note that no samples from $\mathcal{R}$ are available during the few-shot model training.
    \item A \textbf{query set} $\mathcal{Q} = \left\{ x_j \right\}_{j=1}^{Q}$, consisting of unlabeled instances that the model must classify.
\end{itemize}
The goal is to learn a function $f$ that, given a query instance $x_j$ and the support set $\mathcal{S}$, predicts the correct relation $y_j$:

% \begin{equation} \label{eq
% } f(x_j; \mathcal{S}) \rightarrow y_j, \quad y_j \in \mathcal{R}. \end{equation}

% In scenarios incorporating NOTA examples, the model must also determine if the query instance does not correspond to any of the relations in the support set. This adjustment requires modifying Equation \ref{eq
% } as follows:

\begin{equation} \label{eq
} f(x_j; \mathcal{S}) \rightarrow y_j, \quad y_j \in \mathcal{R} \cup {\text{NOTA}}. \end{equation}

A FSRC example is shown in Figure \ref{fig:fsrc-example}.
\subsubsection{Limitations of the $M$-way $K$-shot Evaluation Method}

While the $M$-way $K$-shot evaluation method is commonly employed in few-shot learning, it presents several limitations when applied to FSRC. In FSRC, the key question is: given a new, unseen relation type, what is the expected performance of a model? The current $M$-way $K$-shot evaluation does not provide a straightforward answer to this question. It involves multiple hyperparameters (e.g., $M$, $K$, $Q$, and NOTA percentage)\footnote{The $Q$ parameter refers to the number of query instances used during the evaluation phase of the few-shot learning task.}, leading to a wide range of evaluation configurations. Consequently, results are often tied to specific, arbitrary settings, making it difficult to draw clear conclusions about a model's generalization capabilities. Additionally, the evaluation is influenced by the composition of the query set, introducing bias related to the similarity between the query examples and the support examples.

To address these issues, we propose a simpler and more direct evaluation method that requires the model to determine whether two examples represent the same relation. This alternative evaluation is described in detail and utilized in Section~\ref{sec:expiriemts}. For consistency with existing work, we also employ the traditional $M$-way $K$-shot evaluation when comparing to other datasets. A more thorough comparison between these evaluation methods is provided in Appendix~\ref{app:eval}.

\begin{figure}[h] \centering \includegraphics[scale=0.365]{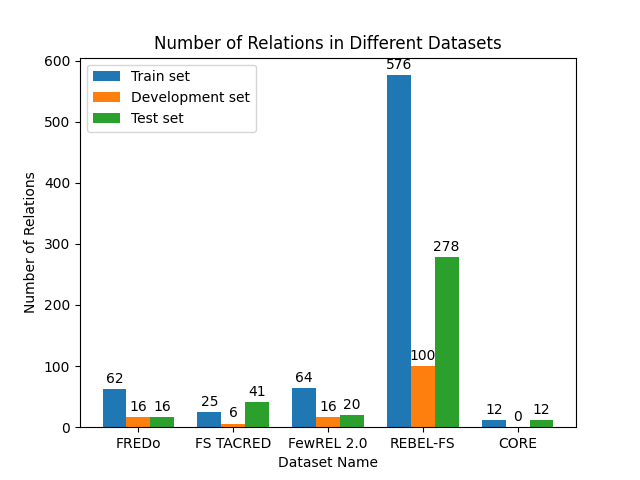} \caption{Comparison of relation type diversity in FSRC benchmarks \citep{popovic-farber-2022-shot, Sabo2021, gao-etal-2019-fewrel, Borchert2023}. REBEL-FS (this work) covers an order of magnitude more relations than existing datasets, enabling a more comprehensive exploration of few-shot generalization.} \label{fig
} \end{figure}

\subsection{Impact of Pretraining Data Diversity}
Recent studies in language model pretraining have highlighted the critical role of data diversity in enhancing model performance. Works by \citet{Gunasekar2023}, \citet{Li2023}, and \citet{Marah2024} demonstrate that models trained on carefully curated and diverse datasets outperform those trained on larger but less diverse datasets, even with fewer parameters. These studies emphasize that factors such as data quality, broad domain coverage, and the avoidance of noisy or biased data are essential for improving model robustness and generalization, rather than merely increasing the volume of training data.

While these findings stem from large-scale language models, they offer valuable insights applicable to few-shot learning tasks beyond FSRC. In contrast to the extensive resources required for pretraining LLMs, few-shot settings provide a more focused and manageable context to investigate the impact of data diversity. By constraining our experiments to FSRC, we can rigorously test the effects of relational diversity on model performance without the computational demands of full-scale LLM training.

\subsection{Large Language Models and FSRC Relation Classification}

LLMs, ranging from 7 billion to over 100 billion parameters, have demonstrated strong capabilities in zero-shot and few-shot learning scenarios due to their extensive training on diverse datasets comprising billions of tokens. These models have shown potential in various NLP tasks, including relation classification \cite{chen2023, hadi2024, ozyurt2024documentlevelincontextfewshotrelation}.

However, the relevance of LLMs to our study is limited for several reasons:

\textbf{Marginal Performance Gains}: While LLMs excel in numerous NLP tasks, their improvements in FSRC tasks are often marginal \cite{li2024llmrelationclassifierdocumentlevel, ozyurt2024documentlevelincontextfewshotrelation}. The complex nature of these models does not significantly enhance relation classification beyond what is achievable with more conventional approaches.

\textbf{Practical and Experimental Constraints}: The large-scale nature of LLMs, combined with the substantial computational costs and infrastructure requirements for training and fine-tuning, pose significant challenges for their deployment in large-scale applications or on resource-constrained devices. This is particularly relevant for tasks that typically operates on vast corpora like FSRC. 

% \subsection{Other Diversity-Related Works}
% \hl{Finish}

\section{The Diversity Hypothesis} \label{sec:diversity}

% \begin{table}
% \centering
% \resizebox{\columnwidth}{!}{%
% \begin{tabular}{lccc}
% \hline
% \textbf{Dataset Name} & \textbf{\# Relation Train}& \textbf{\# Relation Dev}& \textbf{\# Relation Test}\\
% \hline
% FREDo \citep{popovic-farber-2022-shot} & 62 & 16 & 16\\
% FS TACRED \citep{Sabo2021} & 25 & 6 & 41\\
% FewREL 2.0 \citep{gao-etal-2019-fewrel} & 64 & 16 & 20\\
% % REBEL-FS (Ours) &576 & 100 & 278\\
% \hline
% This work &576 & 100 & 278\\
% \hline
% \end{tabular}
% }
% \caption{Number of Relation Types in FSRC datasets.}
% \label{tab:rel_data}
% \end{table}
The primary hypothesis put forth in this paper is that in order to effectively generalize to novel relation types, a diverse dataset is necessary. Furthermore, it is posited that the model's capacity for generalization is more heavily dependent on diversity rather than quantity.

Specifically, we observe that different relations are expressed differently in natural language, through variations in vocabulary and in syntactic configuration, and that these differences are significantly larger across different relation types than within a given relation type: in a small set, the degree of overlap between the characteristics of two arbitrary relations is likely to be low. To effectively recognize new relations, the model must be trained on a diverse dataset that covers a wide range of possible relations between entities.\footnote{An analogy can be drawn to a ``spanning set'' of relations that covers the entire space of possible semantic relations between entities.}

As an example\footnote{The example provided here focuses on entity types in the relations and is intended to serve as an intuition for the primary hypothesis. This is a simplified representation of the classification process.}, consider the relations ``place of birth'' and ``date of birth''. The object of the first relation is ``location'' whereas the object of the second relation is ``time''. By training exclusively on ``the place of birth'' relation, the model may not learn the concept of ``time'' that is necessary to classify ``date of birth'' relations correctly. On the other hand, both relations have a subject of the type person''. Including the first relation in the training, data provides some information that is relevant to the second relation, but this information is incomplete as it does not contain the concept of ``time'' which is necessary for correctly classifying ``date of birth'' cases.

Another example is the ``position held by a person'' relation (e.g. ``Microsoft's CEO Nadella''), which is primarily expressed by a syntactic configuration and is not captured by a restricted set of ``trigger words'', in contrast to many other relations. 
As a final example, consider the relation ``treatment treats condition'', and the need to recognize also constructions such as ``treatment treats patients with condition'', and ``treatment treats patients admitted to hospital with condition''. Such structural variations manifest in different ways for different relations.

By exposing models to a wider range of relation types during training, we aim to equip them with the ability to discern subtler linguistic patterns and generalize more effectively to unseen relations, ultimately leading to improved performance in challenging, real-world FSRC scenarios.

\section{Constructing a Diverse Few-Shot Dataset} \label{sec:comp-datasets}

Existing FSRC datasets often lack sufficient diversity, limiting their ability to accurately assess model generalization. To overcome this limitation, we introduce REBEL-FS, a new few-shot relation classification dataset designed with three key attributes:

\paragraph{1. High Relation Type Diversity:} REBEL-FS is designed to cover a wide range of semantic relationships, ensuring that the dataset includes a large variety of relation types. This diversity is crucial for evaluating a model's ability to generalize to new, unseen relations.

\paragraph{2. Emphasis on Rare Relations:} To reflect real-world challenges, where many important relations are rare, REBEL-FS allocates a significant portion of its evaluation set to rare relations—those with limited representation in typical corpora. This focus helps assess how well models handle less common, specialized relations.

\paragraph{3. Sufficient Representation per Relation:} Each relation type, including the rare ones, is given enough examples to enable meaningful evaluation. This approach ensures that model performance is not overly influenced by small variations in the data.

To achieve these attributes, REBEL-FS is built on the REBEL dataset \cite{Cabot}, a supervised relation classification dataset, which includes 1,146 relation types and over 9 million instances. The following steps were taken to transform REBEL into a few-shot setting:

\paragraph{Data Split:} We split the dataset into training, development, and test sets based on relation frequency. Relations with at least 40 examples were included in the training set, ensuring sufficient representation of common relations. The remaining relations, particularly the rarer ones, were divided between the development and test sets. This approach results in 576 relations for training, 100 for development, and 278 for testing, with an emphasis on evaluating the model's ability to handle rare relations.

\paragraph{Few-Shot Sample Generation:} Each example in REBEL-FS consists of a pair of sentences with marked entity spans. We generate these examples by pairing sentences with the same relation (positive examples) and pairing sentences with different relations (negative examples). To reflect real-world scenarios, where negative examples are more prevalent, we adjust the ratio of positive to negative examples, as detailed in Section \ref{sec:expiriemts}.

\begin{figure}[h]
    \centering
    \includegraphics[width=0.95\columnwidth]{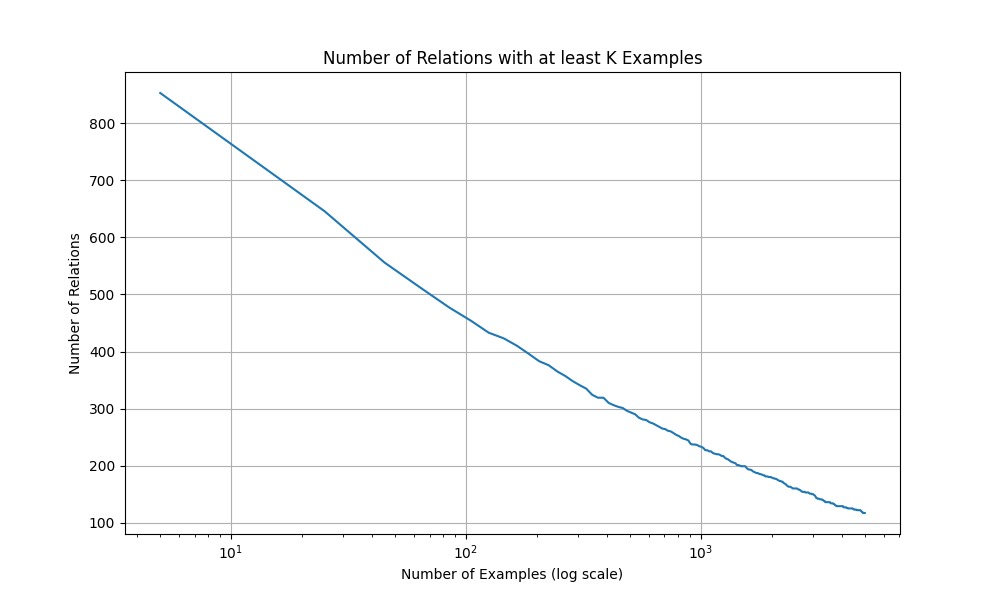}
    \caption{Number of relations with at least K examples in the REBEL-FS dataset. Note that the x-axis is in the log-scale.}
    \label{fig:relation-distribution}
\end{figure}

In summary, REBEL-FS is a diverse and realistic dataset for few-shot relation classification. As shown in Figure \ref{fig:relation-distribution}, REBEL-FS offers a broad distribution of relations, making it a robust benchmark for evaluating model generalization across both common and rare relations.

\section{Experiments and Results} \label{sec:expiriemts}

To assess the impact of relation type diversity on few-shot relation classification (FSRC) performance, we conduct a series of experiments comparing REBEL-FS to several existing FSRC datasets. Additionally, we explore model performance under various challenging settings, such as high-negative instances and varying amounts of training data.

\subsection{Datasets}

We evaluate our proposed REBEL-FS dataset against widely-used FSRC datasets:

\paragraph{FewRel 2.0} \citep{gao-etal-2019-fewrel} A pioneering FSRC dataset sourced from Wikipedia, containing 100 relation types, with 64 used for training, and including "None Of The Above" (NOTA) examples.

\paragraph{CORE} \citep{Borchert2023} Focused on company relations and business entities, this dataset contains 12 relation types, with 5 used for training.

\paragraph{TACRED-FS} \citep{Sabo2021} Derived from the TACRED dataset, this version is designed for high-NOTA scenarios in FSRC, featuring 41 relation types and 24 for training.

\paragraph{REBEL-FS} Our proposed dataset, constructed from the REBEL dataset, contains 954 relation types, with 576 used for training. We systematically vary the number of relation types (10, 50, 100, 200, 400) to analyze the effect of diversity on model performance.

\subsection{Experimental Setup}
We use two setups for evaluation
\subsubsection{Few-Shot Learning Setup}
To ensure consistency with prior work, we adopt the widely used M-way K-shot learning framework \citep{Sabo2021, Han2018, Gao2019,Borchert2023}. As discussed in Section \ref{sssec:formulation}.

\paragraph{Model} We utilize the BERT-Pair model, which has proven effective in FSRC tasks \cite{Sabo2021, Han2018, Gao2019}. Instances are paired with support examples, and the model computes similarity probabilities to determine whether they share the same relation type.

\paragraph{Significance} In all tables reported results in bold are statistically significant (p value < 0.05), while underlined result are highest, but not statistically significant.

\subsubsection{Siamese Setup}  
To further evaluate the effects of diversity, we adopt a streamlined version of the traditional M-way K-shot framework, eliminating most hyperparameters such as \(M\), \(K\), and \(Q\). In this setup, the model is presented with a support set and a single query sample. The objective is to determine whether the query sample expresses the same relation type as the support set or represents a different relation. This approach simplifies the evaluation process compared to M-way K-shot and aligns more closely with real-world scenarios, where users often have a predefined support set for a given relation and seek an estimate of the model’s performance on similar data. Additionally, this framework allows us to assess our hypothesis in a different context beyond the traditional M-way K-shot, providing a broader evaluation of the diversity hypothesis.

\paragraph{Span Representation} Entity in the spans are marked with special tokens (called markers) before being fed into the model. For example, the sentence "Bill Gates worked at Microsoft." is transformed into: ``<s> Bill Gates </s> worked at <o> Microsoft </o>''. The embeddings of these markers are concatenated to form span representations.

\paragraph{Training Details} We use siamese network \cite{Schroff2015, Reimers2020, cohen-etal-2022-mcphrasy} to generate representations for each sentence and calculate the cosine similarity between them. The training objective is to minimize the L2 norm between samples from the same relation and increase it in other cases. More details on the training configuration are provided in Appendix \ref{app:exp-setting}.

\paragraph{Evaluation} In this setup, we evaluate the model by presenting it with two samples, each consisting of a sentence with marked entities indicating the participating entities in the relation, as described in the previous section. The task for the model is to determine whether the two sentences express the same relation or different ones. This evaluation approach is significantly simpler and involves fewer hyperparameters compared to the M-way K-shot method. Furthermore, the results, reported in terms of F1 score, provide a more intuitive and realistic estimate of the model's performance, offering users a clearer understanding of how the model is expected to behave in practical, real-world scenarios.

\subsection{Impact of Relation Diversity on FSRC Generalization} \label{ssec:exp1}

\begin{figure}[h!]
    \centering
    \begin{subfigure}[b]{1.1\columnwidth}
        \centering
        \includegraphics[width=\columnwidth]{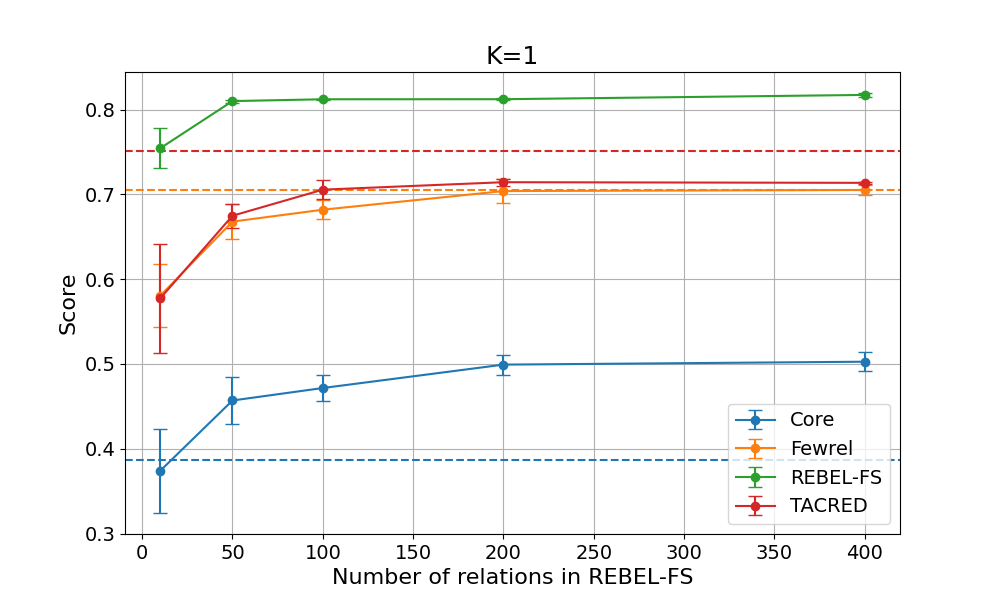}
        \caption{\( K = 1 \)}
        \label{fig:image1}
    \end{subfigure}
    \hfill
    \begin{subfigure}[b]{1.1\columnwidth}
        \centering
        \includegraphics[width=\columnwidth]{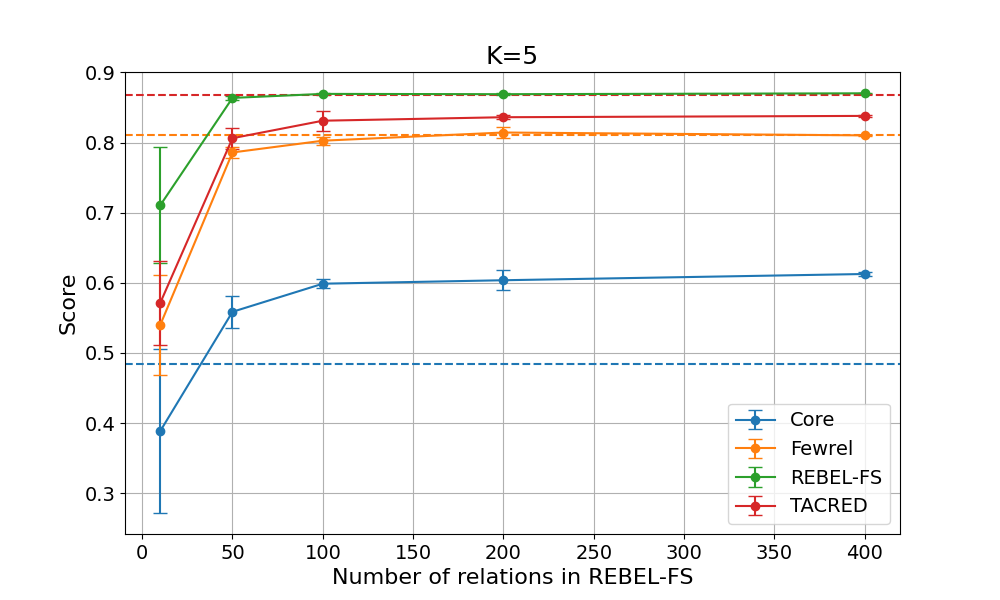}
        \caption{\( K = 5 \)}
        \label{fig:image2}
    \end{subfigure}
    \caption{Performance (accuracy) of REBEL-FS on different test sets based on the number of relations in the training set, with \( K = 1 \) and \( K = 5 \). The dashed lines represent the performance of models trained on the original training data of the datasets.}
    \label{fig:combined}
\end{figure}

We evaluate the generalization ability of REBEL-FS trained models that were trained on varying numbers of relation types (10, 50, 100, 200, 400) across multiple test sets, including FewRel, CORE, and TACRED-FS. To ensure robustness, each model is trained and evaluated three times, with the results averaged across runs (Figure \ref{fig:combined}).

    \paragraph{Diversity Improves Performance:} As the number of relation types increases in REBEL-FS, performance improves across all test sets, showing the positive effects of diversity on model accuracy.
    
     \paragraph{Competitive with FewRel and CORE:} REBEL-FS models trained on 200+ relations either match or surpass the performance of models trained on FewRel and CORE. Notably, REBEL-FS achieves this despite being built from silver-standard data, which is automatically generated and typically noisier than human-annotated data. This show that even with data of lower quality, diversity plays a critical role in achieving competitive performance.

    \paragraph{Lower Performance on TACRED:} The REBEL-FS models performs less effectively on TACRED-FS compared to model that were trained on TACRED-FS, likely due to the unique domain and stylistic differences in newswire text found in TACRED. This demonstrates the limitations of training on datasets with a narrower distribution, where domain-specific characteristics can challenge generalization. Specifically, models trained on datasets like REBEL-FS, which are derived from diverse, open-domain sources such as Wikipedia, may struggle to capture the specific linguistic patterns and entity relations found in more specialized domains like newswire. Such domain-specific nuances, including formal structure, different named entities, and unique relational dependencies, are underrepresented in open-domain data. As a result, the generalization capability of models diminishes when faced with this shift in domain and distribution.

\subsection{Generalization Across Datasets}

We further investigate the generalization capabilities of REBEL-FS by training models on one dataset and evaluating them on another. As shown in Table \ref{tab:K1K5}, REBEL-FS demonstrates competitive performance across Wikipedia-based datasets (FewRel and CORE), but generalizes less effectively to TACRED-FS, reinforcing the importance of domain-specific training.

\begin{table}[h!]
  \centering
  \resizebox{\columnwidth}{!}{
  \begin{tabular}{ccccc}
    \hline
    \textbf{Test / Train} & \textbf{CORE} & \textbf{FewRel} & \textbf{REBEL-FS} & \textbf{TACRED} \\
    \hline
    \multicolumn{5}{c}{\textbf{K = 1}} \\
    \hline
    CORE & 0.39 & 0.54 & 0.69 & 0.43 \\
    FewRel & 0.48 & \textbf{0.71} & 0.80 & 0.73 \\
    REBEL (400) & \textbf{0.50} & \textbf{0.71} & \textbf{0.82} & 0.71 \\
    TACRED & 0.45 & 0.66 & 0.77 & \textbf{0.75} \\
    \hline
    \multicolumn{5}{c}{\textbf{K = 5}} \\
    \hline
    CORE & 0.48 & 0.69 & 0.81 & 0.57 \\
    FewRel & 0.56 & \textbf{0.81} & 0.86 & 0.86 \\
    REBEL (400) & \textbf{0.61} & \textbf{0.81} & \underline{0.87} & 0.84 \\
    TACRED & 0.51 & \textbf{0.81} & 0.85 & \underline{0.87} \\
    \hline
  \end{tabular}
  }
  \caption{\label{tab:K1K5} Performance metrics for models trained on various datasets and evaluated on test sets of other FSRC datasets.}
\end{table}

\subsection{Ablation Study}

\begin{table*}[h]
\centering
\resizebox{1.8\columnwidth}{!}{%
\begin{tabular}{rr|ccc|ccc|ccc}
\hline
%\textbf{\# of rel types}
\textbf{\# rel} & \textbf{data} & \multicolumn{3}{c|}{\textbf{0.5 negatives}} & \multicolumn{3}{c|}{\textbf{0.9 negatives}} & \multicolumn{3}{c}{\textbf{0.99 negatives}} \\
%\hline
\textbf{types} & \textbf{size} & F1 & P & R & F1 & P & R & F1 & P & R \\
\hline
\textbf{29} & 100K &80.9 & 76.98 & 85.24 & 41.12& 27.24 & 83.87&6.39& 3.32&83.82 \\
\textbf{79} & 100K &85.8 &84.07 &87.61 &52.68& 37.9& 86.33& 10.19& 5.41& 86.09 \\
\textbf{233} & 100K &88.96 &88.19 & 89.74 &58.54&43.56& 89.21&11.9&6.38& 89.12 \\ 
\textbf{461} &100K &90.17& \underline{91.85} & 88.55 &66.48& \textbf{53.43} & 87.94&\textbf{16.42}& \textbf{9.05}& 88.22 \\
\textbf{576} & 100K &\textbf{91.36}& 91.23& \textbf{91.49}& \underline{66.52} & 55.38 & \textbf{91.11} &14.22& 7.71& \textbf{91.25} \\
\hline
\hline
\textbf{576} & 1K & 90.75 & 93.97 & 87.75 & 65.21 & 55.28 & 86.38 &  15.19 & 8.33 & 86.28\\
\textbf{576} & 10K & 91.25 & 90.24 & 92.29 & 63.43 & 48.48 & 91.72 & 14.34 & 7.78 & 91.67\\
\hline
\end{tabular}
}
\caption{Model performance based on the number of different relation types, NOTA percentage, and sample size. Models were evaluated on a test set that contained 50\%, 10\%, and 1\% of positive samples.}
\label{tab:rel_results}
\end{table*}

Using the Siamese setup, we conduct experiments varying the number of relation types (29 to 576), NOTA percentage, and train sample size. We create groups of relations s.t. each group contains all relations having at least $M$ instances, where $M \in \{5000, 1000, 500, 100, 40\}$ resulting in 29, 79, 233, 461, and 576 different relation types and each train set contained 50\% positive pairs (where both samples correspond to the same relation) and 50\% negative (where both samples correspond to the different relations). Results show that increasing relation diversity leads to significant gains in F1 score  across all variables, confirming our hypothesis that diversity aids generalization and overall model performance (Table \ref{tab:rel_results}).

% \paragraph{High Diversity Leads to High generalization}
% Like in Section \ref{ssec:exp1}, The results in Table \ref{tab:rel_results} indicates that the performance of the model increase in relation to to the diversity of the dataset.

\paragraph{High-Negative Ratios}
% We test models in high-negative settings (90\% and 99\% negative examples) to simulate realistic FSRC scenarios. Models trained with more relation types demonstrate better performance, with F1 gains of up to 66.5\% in 90\% negative scenarios and 157\% in 99\% negative scenarios.

We evaluate the models in high-negative settings, specifically with 90\% and 99\% negative examples, to simulate realistic FSRC scenarios. The results demonstrate a clear benefit from increasing the diversity of relation types in the training data. In the 90\% negative setting, models trained on 576 relation types achieve an F1 score of 66.52, a significant improvement compared to the 41.12 F1 score of the model trained on only 29 relation types, marking a 61.8\% gain. Similarly, in the 99\% negative scenario, the model trained on 576 relation types achieves an F1 score of 14.22, substantially outperforming the 6.39 F1 score achieved by the model trained on 29 relation types, reflecting a remarkable 122.5\% increase. The substantial F1 improvements, particularly in high-negative settings, further validate the hypothesis that relation type diversity is a crucial factor for enhancing model generalization and robustness in real-world FSRC tasks.

\paragraph{Effect of Training Data Size}
Interestingly, we find that models trained with a smaller dataset (1\% or 10\% of the total data) perform comparably to models trained on the full dataset, provided there is sufficient relation type diversity. This suggests that dataset size can be reduced significantly without sacrificing performance, as long as diverse relation types are included.

Interestingly, we observe that models trained on significantly smaller datasets (1\% and 10\% of the baseline dataset) exhibit competitive performance compared to those trained on the full dataset, provided that the relation type diversity is maintained. For instance, the model trained with only 1K examples (1\% of the data) on 576 relation types achieves an F1 score of 90.75 in the 50\% negative scenario, which is remarkably close to the F1 score of 91.36 achieved by the model trained on the full 100K dataset. Similarly, in more challenging settings, such as with 90\% and 99\% negatives, the smaller dataset shows only minor drops in performance, with F1 scores of 65.21 and 15.19, respectively, compared to 66.52 and 14.22 for the full dataset. These results demonstrate that dataset size can be significantly reduced without sacrificing much in terms of performance, as long as relation type diversity remains high. This suggests that models can be trained efficiently even with reduced data.

\subsection{Learning Curves and Overfitting}
% We monitor training performance (on the development set) over multiple epochs to assess overfitting. Models trained with fewer relation types (29 and 79) show signs of overfitting early in training, while models with 233+ relation types generalize better, displaying more stable learning curves (Figure \ref{fig:overfiting}).

\begin{figure}
\includegraphics[scale=0.37]{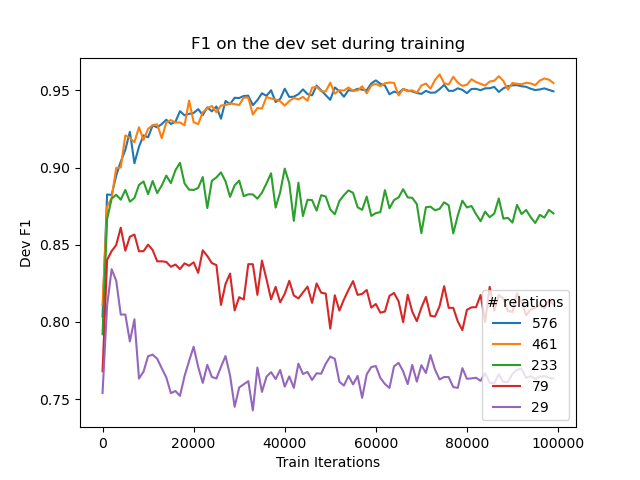}
\caption{F1 as a function of training steps during training.} \label{fig:overfiting}
\end{figure}

We investigate how the diversity of relation types affects the model’s propensity to overfit. To monitor this, we evaluate the models every 1000 training steps over the course of 4 epochs. The results, shown in Figure \ref{fig:overfiting}, reveal several notable trends. First, all models display a strong baseline performance, likely attributable to the pretrained model. Second, models trained on only 29 and 79 relation types show minimal improvement over this baseline, and they exhibit clear signs of overfitting as early as the first epoch. In contrast, the model trained on 233 relation types begins to overfit after approximately two epochs. Finally, models trained on a more diverse set of 461 and 576 relation types show no observable overfitting throughout the training process. These findings suggest that high relation diversity might make training more stable and mitigate overfitting.

\section{Conclusion} \label{sec:conclusion}
This work has highlighted the importance of diversity in the training data for few-shot relation classification. We have demonstrated that increasing the number of relation types in the training data leads to significant improvements in model performance, especially in highly negative scenarios. Our results show that even a small amount of diverse data can achieve performance comparable to that of a large dataset, challenging the notion that large amounts of data are necessary for high performance in natural language understanding tasks. Additionally, we have shown that diversifying the training data can mitigate overfitting and improve generalization. These findings have important implications for the development of future datasets and training strategies in the field of relation classification and the ability of language model to generalize.
% \subsection{Few rel 2.0}
% To show that our diversity hypothesis is algorithm invariant, and is not dependent on our distantly supervised dataset, we test our assumption on the few rel 2.0 datasets \cite{gao-etal-2019-fewrel}. We do this by using MNAV \cite{Sabo2021}, and by replacing the training dataset with our training data, where once again we fix the train set size and change the number of training relation types.

\section*{Limitations}
%Additionally, 
While our results suggest that diversifying the training data is crucial for achieving high performance in FSRC, it is important to note that there may be other factors that contribute to the poor performance of current FSRC datasets, such as the quality of the annotations and the difficulty of the task itself. Despite these limitations, our work highlights the importance of diversity in training data for achieving high performance in FSRC and provides a valuable perspective for future research in this area.
\section*{Ethics Statement}
To the best of our judgement, this work does not raise any significant ethical concerns.

\section*{Acknowledgements}
This project has received funding from the European Research Council (ERC) under the European Union's Horizon 2020 research and innovation program, grant agreement No. 802774 (iEXTRACT).

\bibliography{anthology,custom}
\bibliographystyle{acl_natbib}

\clearpage
\newpage
\section*{Appendix}
\appendix

\section{Experimental Setting}
\label{app:exp-setting}

\begin{table*}[h!]
\centering

\resizebox{2\columnwidth}{!}{%
\begin{tabular}{crr}
\hline
\textbf{Parameter} & \textbf{RECESS-FS}& \textbf{m-way k-shot}\\
\hline
\textbf{Batch size} & 4 & 64 \\
\textbf{training steps} & 4 epocs & 500 steps\\
\textbf{Learning rate} & $2\cdot10^{-5}$ & $2\cdot10^{-5}$\\
\textbf{Weight decay} & 0.1 & 0\\
\textbf{GPUs used} & 1 & 1\\
\textbf{GPU type} & Nvidia 2080TI & Nvidia 3090 \\
\textbf{pretrained model} & SpanBERT \cite{Joshi2019} & bert \cite{Devlin2018} \\
\textbf{Optimizer} & Adam & Adam\\
\textbf{Train time} & 3-4 hour & 0.5 hour (5-1), 2 hours (5-5) \\
\hline
\end{tabular}
}
\caption{Training parameters.}
\label{tab:train_statistics}
\end{table*}

Training parameters for all experiments are available in \cref{tab:train_statistics}. %\shauli{need to add details about architecture, training time, optimizers, stopping criterion, GPUs} 

\section{Evaluation} \label{app:eval}
To effectively evaluate the impact of relation type diversity on few-shot relation classification (FSRC) performance, a robust and informative evaluation methodology is essential. This section introduces our proposed evaluation scheme, contrasting it with the traditional M-way K-shot approach (even with NOTA incorporation) and highlighting its advantages for assessing generalization ability in realistic FSRC scenarios.

\paragraph{M-way K-shot Evaluation (with NOTA).}
While the standard M-way K-shot paradigm doesn't inherently include NOTA, it can be adapted to do so. In this variant, one of the "M" relations represents the NOTA class. The support set contains K examples for each of the M-1 target relations and K examples of the NOTA class. The model then classifies Q query instances into one of these M classes (M-1 relations + NOTA). However, even with NOTA incorporated, this approach still presents challenges:
\begin{itemize}
    \item \textbf{Indirect NOTA Control:} While the proportion of NOTA instances in the evaluation set can be adjusted, it is not directly comparable to real-world NOTA percentages, which are often significantly higher than the proportion of any individual target relation.
    \item \textbf{Convoluted Difficulty Control:} The interplay between M, K, and the desired NOTA percentage can make it difficult to precisely control the evaluation's difficulty and align it with real-world scenarios or existing benchmarks.
\end{itemize}

\paragraph{Proposed Evaluation: Relation-Specific Classification.}
Our proposed approach directly addresses these limitations. Given a support set consisting of one or more examples demonstrating a single target relation, the model must determine whether a given target sample also exhibits this relation. This approach offers several key benefits:
\begin{itemize}
\item \textbf{Realistic and Controllable Difficulty:} Our approach allows precise control over the NOTA percentage, directly simulating the high-negative scenarios common in real-world FSRC. This also enables fine-grained difficulty control and alignment with estimated NOTA percentages in existing datasets for more meaningful comparisons.
\item \textbf{Intuitive and Informative Evaluation:} The evaluation closely mirrors real-world relation extraction tasks and provides a directly interpretable F1 score that reflects the model's ability to accurately retrieve relevant relations.
\item \textbf{Simplicity and Transparency:} Our method is straightforward to implement, requiring only a binary classification for each target sample. This simplicity promotes transparency and ease of analysis compared to more complex evaluation schemes.
\end{itemize}
Our relation-specific evaluation offers a more realistic, controllable, and informative assessment of FSRC models, allowing us to directly investigate the impact of relation type diversity on generalization ability, particularly in challenging, high-NOTA scenarios. The remainder of this section details our experimental setup and results using this evaluation method.
In our experiments, we utilize our proposed relation-specific evaluation to assess model performance on REBEL-FS. For consistency and comparability with prior work, we employ the M-way K-shot evaluation (with NOTA) when evaluating models on existing FSRC datasets. However, we believe our proposed metric provides a more robust and informative measure for future FSRC research and encourages its adoption in subsequent benchmarks.

\end{document}